\documentclass[letterpaper]{article} 
\usepackage[preprint]{aaai2027}  
\usepackage{times}  
\usepackage{helvet}  
\usepackage{courier}  
\usepackage[hyphens]{url}  
\usepackage{graphicx} 
\usepackage{natbib}  
\usepackage{caption} 
\usepackage{subfig}
\usepackage{algorithm}
\usepackage{algorithmic}
\usepackage{newtxmath}
\usepackage{mathtools}
\usepackage{multirow}
\usepackage{array}
\usepackage{booktabs}
\usepackage{newfloat}
\usepackage{listings}
\DeclareCaptionStyle{ruled}{labelfont=normalfont,labelsep=colon,strut=off} 
\floatstyle{ruled}
\newfloat{listing}{tb}{lst}{}
\floatname{listing}{Listing}
\usepackage{xcolor}%

\usepackage{booktabs}
\usepackage{multirow}
\usepackage{graphicx}
\usepackage[table]{xcolor}
\usepackage{tabularx}
\usepackage{array}

\title{Task-Anchored Representation Shaping for Pre-Trained \\ Model-Based Continual Learning}
\author{
    Zhiming Xu\textsuperscript{\rm 1,2},
    Huiyu Yi\textsuperscript{\rm 1,2},
    Zhen-Hao Xie\textsuperscript{\rm 1,2},
    Baile Xu\textsuperscript{\rm 1,2},
    Furao Shen\textsuperscript{\rm 1,2,\textdagger},
    Jian Zhao\textsuperscript{\rm 4},
    Suorong Yang\textsuperscript{\rm 1,3}
}
\affiliations{}

\makeatletter
\newcommand\blfootnote[1]{%
  \begingroup
  \renewcommand\thefootnote{}%
  \footnote{#1}%
  \addtocounter{footnote}{-1}%
  \endgroup
}
\makeatother

\usepackage{bibentry}

\begin{document}

\maketitle
\blfootnote{
\textsuperscript{\rm 1}National Key Laboratory for Novel Software Technology, Nanjing University, Nanjing, China.
\textsuperscript{\rm 2}School of Artificial Intelligence, Nanjing University, Nanjing, China.
\textsuperscript{\rm 3}School of Computer Science, Nanjing University, Nanjing, China.
\textsuperscript{\rm 4}School of Electronic Science and Engineering, Nanjing University, China.
\textsuperscript{\textdagger}Correspondence to: \texttt{frshen@nju.edu.cn}
}

\begin{abstract}
Pre-trained models (PTMs) provide a strong foundation for continual learning by offering stable representations that facilitate lightweight adaptation to new tasks. However, adapting well to each task does not ensure reliable inference over all learned tasks. Since task boundaries are often artificial and semantically entangled, an input from an unknown task can remain ambiguous even with strong PTM features, making cross-task prediction a key bottleneck.
We propose Task-Anchored Inference Latent Shaping (TAILS), a lightweight post-PTM module that can be integrated into diverse continual learners and optimized through a decoupled step. TAILS uses fixed task anchors as persistent references to accumulated knowledge. It interprets each sample's feature representation relative to these references, then composes relevant evidence across tasks into latent recall. Rather than selecting a task-specific path or adjusting classifier outputs, TAILS uses latent recall to directly correct the feature representation before prediction. It therefore resolves cross-task ambiguity at the representation level, while leaving the original PTM, method-specific modules, and classifier unchanged.
Extensive experiments across multiple PTM-based continual learning paradigms show that TAILS can improve classification and task-inference performance with modest parameter overhead and negligible inference cost.

\end{abstract}
\section{Introduction}
Modern deep models are increasingly deployed in non-stationary environments, where data arrive sequentially, and their distributions evolve over time~\cite{gomes2017survey,ye2019learning,chen2021large,chen2022learning,yang2024entaugment}. To remain effective in such environments, models must continually incorporate new knowledge without erasing what has already been learned, which is the central objective of continual learning (CL)~\cite{liangloss,yang2025supervised}. 
However, sequential updates can cause catastrophic forgetting~\cite{de2021continual}, whereby learning new data overwrites previously acquired knowledge. Although conventional CL methods alleviate this problem~\cite{zheng2025task,aghasanli2025prototype,li2025re,nori2025federated}, they still learn representations primarily from limited incremental, task-specific data, restricting their ability to maintain a stable and discriminative feature space across tasks. Pre-trained models (PTMs) offer a stronger foundation for CL by supplying representations learned from large-scale data~\cite{wang2022dualprompt,zhou2025revisiting,wang2025tuna,sun2025mos}. Rather than learning representations from scratch on scarce incremental data, PTM-based methods adapt to incoming tasks through lightweight components while keeping the backbone frozen or lightly updated, yielding stable and discriminative features throughout continual learning.

\begin{figure}[!t]
  \centering
  \subfloat[Task-Inference Bottleneck]{
    \includegraphics[height=1.38in,keepaspectratio]{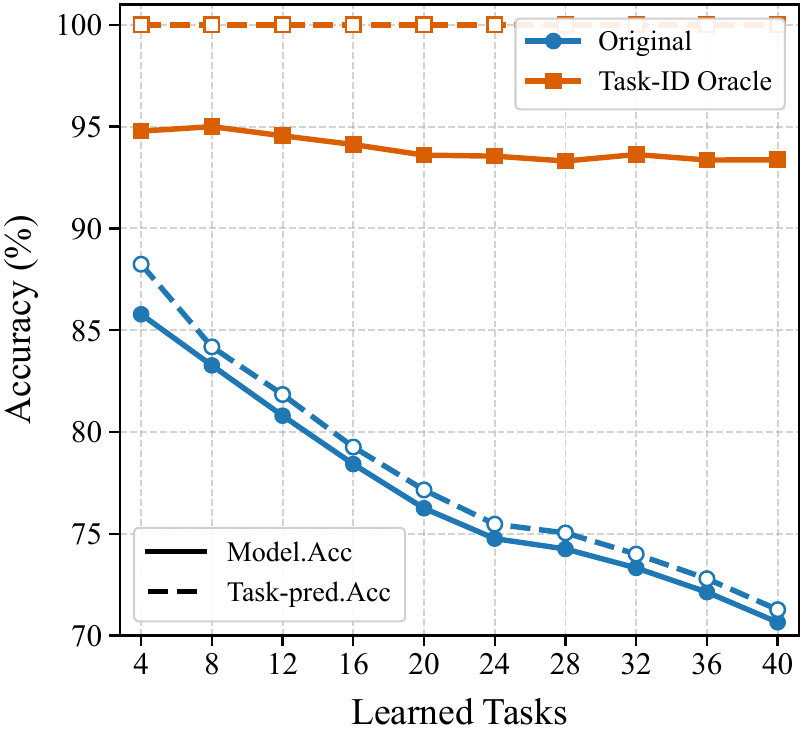}
    \label{fig:fig1a}
    }
  \subfloat[Recall-Guided Reshaping]{
    \includegraphics[height=1.38in,keepaspectratio]{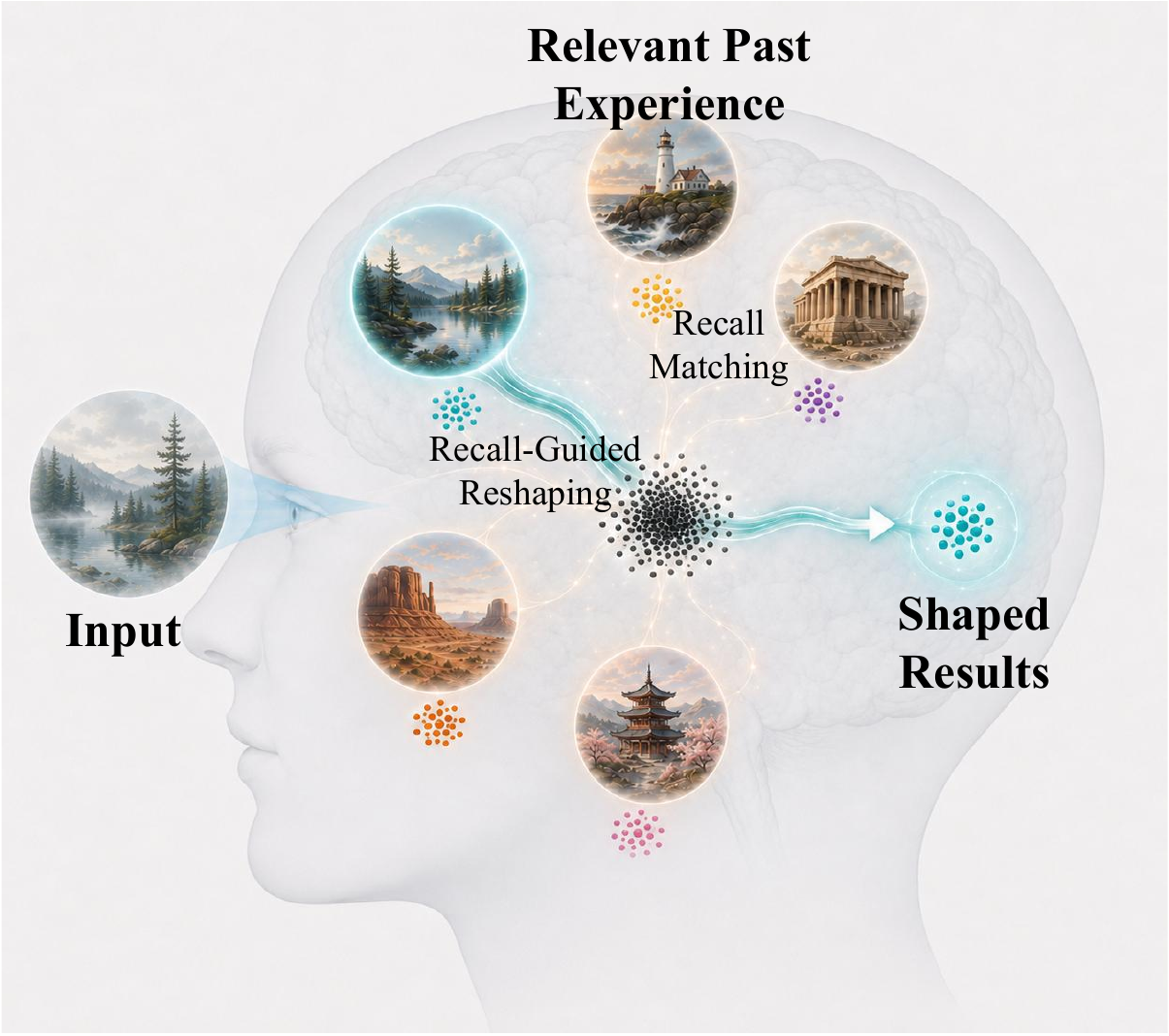}
    \label{fig:fig1b}
    }  
  \caption{Motivation of TAILS.
\textbf{Left}: Accuracy under the original and task-ID-oracle settings throughout continual learning.
\textbf{Right}: For an ambiguous input, TAILS interprets its representation against accumulated task structures, composes sample-specific latent recall, and uses it to condition a feature correction before prediction by the original classifier.}
  \label{fig:fig1}
\end{figure}

Despite these advantages, data arriving at each learning stage are conventionally organized as distinct tasks~\cite{zhou2024continual}, whereas inference must be performed over all accumulated tasks without task identity. As powerful PTMs and task-wise adaptation have substantially reduced intra-task prediction errors, the primary challenge has shifted from learning individual tasks to identifying and exploiting relevant cross-task knowledge, making this identification and use increasingly important.
For example, in continual classification~\cite{zhou2024class}, classes are typically assigned to tasks in a random order so that the resulting task boundaries may exhibit little meaningful semantic separation. Consequently, inputs from different tasks may remain highly overlapping in the feature space, making task inference inherently ambiguous. 
This ambiguity becomes increasingly severe as more tasks are accumulated, since the learner must distinguish among a growing number of semantically entangled task distributions.
Existing methods mainly address this problem through task selection~\cite{wang2022dualprompt,xu2025dual,sun2025mos} or model ensembling~\cite{zhou2024ease,wang2025tuna}. These strategies organize task-specific components through routing or aggregation, but their effectiveness depends heavily on accurately estimating task relevance. When task boundaries lack clear semantic separation, inaccurate task inference inevitably leads to selecting or weighting inappropriate task-specific components, while leaving the underlying feature ambiguity unresolved.
Fig.~\ref{fig:fig1a} illustrates that under a Task-ID Oracle setting, where ground-truth task identity is supplied at inference, accuracy for existing PTM-based CL methods remains nearly flat as tasks accumulate; without this oracle, task-prediction accuracy drops sharply and converges with overall accuracy in later tasks. 
This convergence suggests that existing methods already perform reliably once the correct task is identified, whereas errors in task inference increasingly dominate final prediction performance, making task inference a major bottleneck for continual learning.
This naturally raises a new and pressing question for PTM-based CL: \textit{how can a PTM-based CL learner effectively acquire and preserve task-level structures for reliable cross-task inference?}

To address the cross-task inference bottleneck, we internalize accumulated task structures into the current representation before prediction. Specifically, we propose Task-Anchored Inference Latent Shaping (TAILS), a lightweight plug-and-play module inserted after the PTM feature encoder and before the classifier. Unlike selection and ensembling, TAILS does not route samples to a discrete task-specific component; instead, it composes a soft, input-conditioned summary of evidence across all learned tasks and uses it to correct the feature itself, leaving the original classifier and prediction mechanism unchanged. Concretely, given a PTM-derived feature, TAILS forms a compact query, matches it against task slots built from accumulated task anchors, composes the retrieved evidence into a sample-specific latent recall, and uses this recall to condition a correction of the feature before classification (Fig.~\ref{fig:fig1b}).  TAILS is shared across all learned tasks and trained in a decoupled step after the base learner's standard training completes, so it augments inference without altering or risking the base learner's original training pipeline. 
To keep this recall process stable as new tasks arrive, we construct fixed task anchors from class prototypes and support features, which serve as compact, persistent references both for preserving the recall space during optimization and for retrieving accumulated evidence at inference. 
Because TAILS is a plug-in, post-hoc module rather than a replacement for any existing component, it can be attached to a wide range of PTM-based continual learners with modest parameter overhead and negligible inference cost. It consistently elevates diverse state-of-the-art baselines such as TUNA and CL-LoRA by up to 5.05 percentage points while incurring as little as 3.0\% additional storage overhead when added to storage-heavy base learners, highlighting both its efficiency and effectiveness.
The main contributions are summarized as follows: \textbf{(1)} We propose TAILS, a plug-in post-PTM module that performs assisted inference with task-anchored recall, enabling base learners to shape representations without modifying their original prediction mechanisms. \textbf{(2)} We construct task anchors from class prototypes and support features. These anchors provide compact references for preserving and accessing accumulated task structures during optimization and inference. \textbf{(3)} We conduct extensive experiments across diverse PTM-based CL paradigms, showing that TAILS consistently improves performance with modest parameter overhead and negligible inference cost.




\section{Related Work}
\subsection{Continual Learning}
Continual learning (CL) aims to continuously acquire knowledge from non-stationary data streams while preserving previously learned capabilities. A broad body of work has addressed this objective across diverse continual settings through complementary mechanisms~\cite{zhu2025adaptive,zhang2025revisiting,zhou2025dual,liu2025enhancing,gao2026diapr}. Replay-based methods retain past information by storing representative samples or reconstructing compact historical distributions from prototypes and features~\cite{zhu2025adaptive,zhang2025revisiting}. Regularization- and distillation-based approaches constrain parameter updates or model responses to remain consistent with previously learned functions~\cite{gao2025fairness,zang2026topology}. Prototype- and feature-based methods preserve class-level statistics, refine historical prototypes, or compensate for representation drift across increments~\cite{zhu2025adaptive,gao2026diapr}. Architecture- and ensemble-based methods instead isolate or combine knowledge by expanding model capacity, organizing related classes, or aggregating multiple model states~\cite{lai2025order,lee2025tripartite}. 
However, these methods still construct and maintain their knowledge from the limited incremental stream, which constrains the quality and generalizability of the learned feature space. This limitation has motivated the growing adoption of large-scale pre-trained models as a stronger representational foundation for continual learning.

\begin{figure*}[!ht]
  \centering
  \includegraphics[width=4.8in,keepaspectratio]{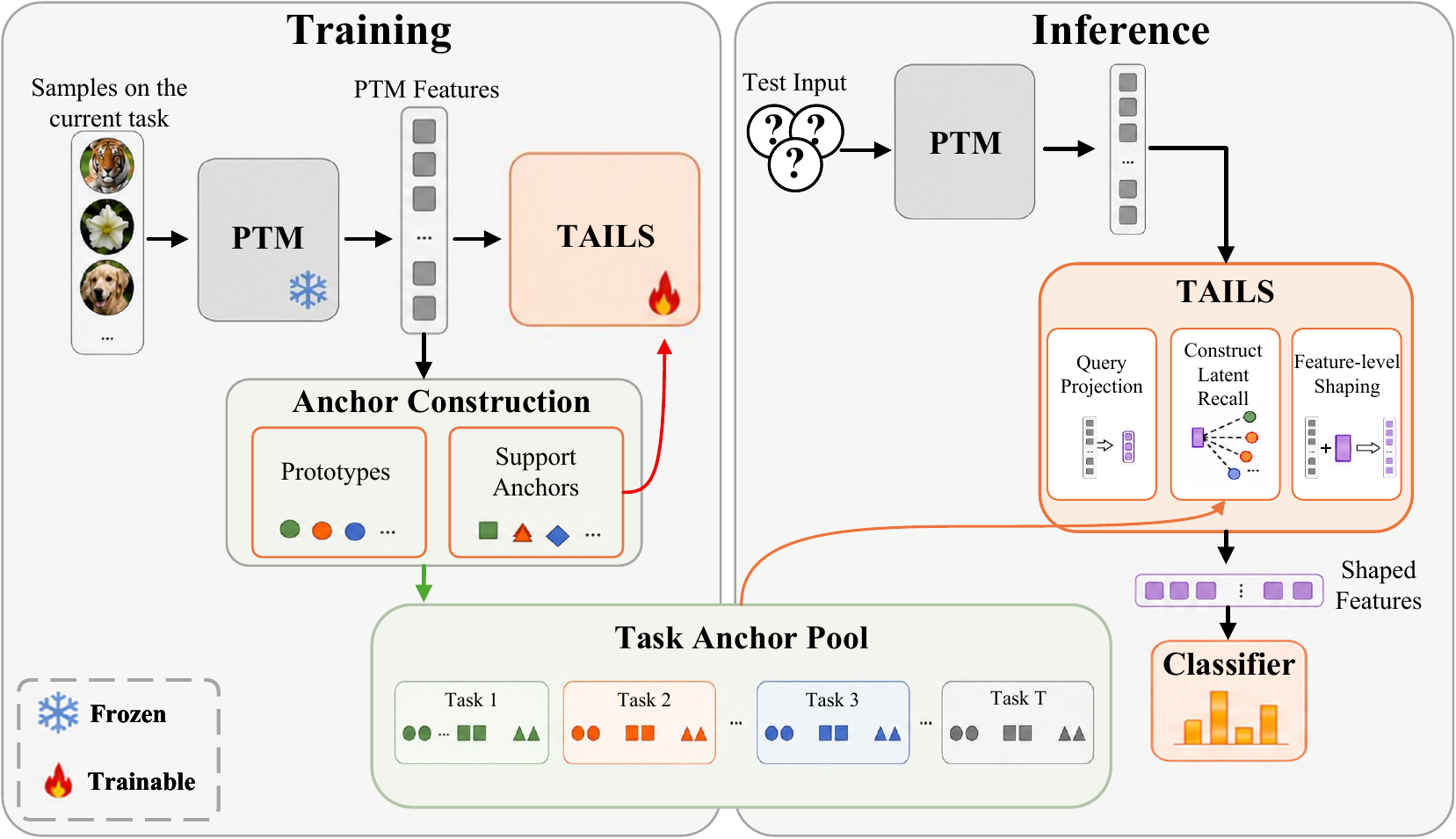}
  \caption{Illustration of TAILS. After each task is learned by the base learner, its class prototypes and features are retained as task anchors. During continual optimization, the accumulated anchor pool maintains a coherent latent-recall space. At inference, a query derived from the current representation addresses the accumulated task structures and composes sample-specific latent recall, which conditions feature shaping before prediction by the original classifier.}
    \label{fig:ours}
\end{figure*}

\subsection{Pre-Trained Model-based CL}

Recent studies have increasingly incorporated pre-trained models into continual learning. Early approaches adapt frozen PTMs with learnable prompts, including L2P~\cite{wang2022learning}, DualPrompt~\cite{wang2022dualprompt}, CODA-Prompt~\cite{smith2023coda}, and APT~\cite{chen2025apt}. Adapter-based methods insert lightweight modules for efficient task adaptation. EASE~\cite{zhou2024ease} builds task-specific adapter subspaces and ensembles their representations, while MOS~\cite{sun2025mos} trains task-specific adapters with improved retrieval to reduce forgetting. SEMA~\cite{wang2025sema} dynamically expands modularized adapters under distribution shifts, whereas TUNA~\cite{wang2025tuna} combines task-specific and universal adapters for specialization and knowledge sharing. In addition, MOS and TUNA preserve class-alignment statistics from previous tasks, which can be viewed as a form of non-sample replay. Such retained statistics mainly support adapter retrieval and classifier calibration, rather than directly resolving sample-specific ambiguity at the representation level before prediction. LoRA-based methods perform adaptation through low-rank updates. SD-LoRA~\cite{wu2025sdlora} decouples LoRA direction and magnitude, while CL-LoRA~\cite{he2025cllora} combines task-shared and task-specific low-rank adapters. Other works revisit PTM-based CIL through prototype learning~\cite{zhou2025revisiting} and analytic classifiers~\cite{yi2026beyond}, further demonstrating the benefits of strong PTM representations for continual recognition.

\section{Preliminaries}
A continual learning (CL) model learns from a sequence of evolving data streams
$D_1, D_2,\cdots,D_t$, where each stream
$D_i=\{(\mathbf{x}_j,y_j)\}_{j=1}^{n_i}$ contains $n_i$ training samples.
For any stage $t$, the current stream $D_t$ is treated as a new task, with its label space denoted by $Y_t$. The model has encountered the data streams
$D_{1:t}=D_1\cup D_2\cup\cdots\cup D_t$ and the accumulated label space
$\mathcal{Y}_{1:t}=Y_1\cup Y_2\cup\cdots\cup Y_t$.
The goal is to learn a CL model
$f_t:\mathcal{X}\rightarrow\mathcal{Y}_{1:t}$ that performs well on all learned data. 
The performance of $f_t$ is evaluated by its classification error on the accumulated test set:
\begin{equation}
    \label{eq:s3e1}
    \operatorname{Err}(f_t)
    =
    \mathbb{E}_{(\mathbf{x},y)\in D^{\mathrm{test}}_{1:t}}
    \left[
        \mathbb{I}\left(f_t(\mathbf{x})\neq y\right)
    \right],
\end{equation}
where $\mathbb{I}(\cdot)$ is the indicator function. We follow the commonly adopted class-incremental continual learning setting, where task IDs are unavailable at inference. The model must make correct predictions among all classes using only the sample. 

This paper focuses on PTM-based CL, where the learner adopts a powerful pre-trained encoder as its feature extractor, therefore providing a stronger foundation for continual adaptation. Given an input sample $\mathbf{x}$, a PTM-based learner first extracts its representation through the pre-trained encoder $\phi_{\mathrm{ptm}}$ and then predicts the label with a classifier $G$:
\begin{equation}
    \mathbf{h}
    =
    \phi_{\mathrm{ptm}}(\mathbf{x}),
    \quad
    \hat{y}
    =
    G(\mathbf{h}).
\end{equation}
To better adapt the PTM to downstream continual tasks, recent PTM-based CL methods usually freeze the PTM backbone and introduce lightweight fine-tuning modules. The PTM-based learner is trained only on the current task data $D_t$ at task $t$; once it proceeds to the next task, previous training images are inaccessible.

\section{Method}
TAILS introduces \emph{task-anchored assisted inference} at the feature level. Unlike task routing, which selects a task-specific path, or output calibration, which adjusts logit scores from the classifier, TAILS performs auxiliary inference directly within the feature space. Specifically, a shared post-PTM module uses the retrieved latent recall to reposition an ambiguous input in the accumulated class space, while leaving the base learner and its original prediction mechanism unchanged, as illustrated in Fig.~\ref{fig:ours}.
In the following, we describe the TAILS module, its training objective, and its integration with different PTM-based learners.

\subsection{Task-Anchored Latent Recall and Shaping}
TAILS is a unified post-PTM module
$\mathcal{F}_{\mathrm{TAILS}}(\mathbf{h};\mathcal{A})$
shared across all learning tasks. It interprets the current representation against task structures retained in the accumulated task anchors, composes relevant historical evidence into latent recall, and internalizes the recalled evidence as a feature correction.
After the base learner finishes task $i$,  we construct an anchor subset $\mathcal{A}_i$ from its post-update sample features. For each class, we retain its class prototype and uniformly sample $M$ support features without replacement, resulting in a class-balanced anchor set. If fewer than $M$ samples are available, all features of that class are retained. The prototypes capture class-level semantics, while the support anchors preserve local feature variations. 
Once constructed, each $\mathcal{A}_i$ remains fixed, and the accumulated task anchor pool $\mathcal{A}=\{\mathcal{A}_1,\ldots,\mathcal{A}_t\}$ provides a persistent reference for cross-task optimization and inference. Since continual tasks may overlap semantically, a representation that is discriminative within its local task can remain ambiguous among classes learned at different tasks. TAILS addresses this ambiguity through soft, input-dependent access to the accumulated anchor structures.
To establish this access, TAILS represents the current input and each task-specific anchor structure in the same compact space. Given a representation $\mathbf{h}\in\mathbb{R}^{d}$ extracted from the input sample, we first derive a compact normalized query $\mathbf{q}\in\mathbb{R}^{d_q}$, where $d_q=64$:
\begin{equation}
    \mathbf{q}
    =
    \mathrm{Norm}
    \left(
        \mathrm{LN}(W_q\mathbf{h})
    \right),
    \label{eq:tails_query}
\end{equation}
where $\mathrm{LN}(\cdot)$ denotes layer normalization \cite{ba2016layer}, and $\mathrm{Norm}(\cdot)$ denotes $\ell_2$ normalization. The query provides an input-conditioned address for accessing the accumulated anchor structures.
Let $\mathbf{A}_i\in\mathbb{R}^{|\mathcal{A}_i|\times d}$ denote the matrix obtained by stacking all anchors in $\mathcal{A}_i$. To make the stored anchors directly comparable with the query $\mathbf{q}$, TAILS projects every anchor of task $i$ from the original feature space into the same $d_q$-dimensional query space:
\begin{equation}
    \overline{\mathbf{A}}_i
    =
    \mathrm{Norm}
    \left(
        \mathrm{LN}
        \left(
            \mathbf{A}_i W_a^{\top}
        \right)
    \right).
\end{equation}
Then TAILS summarizes the embedded anchor structure into $K$ task slots:
\begin{equation}
    \mathbf{Z}_i
    =
    \mathrm{Norm}
    \left[
        \mathrm{softmax}
        \left(
            \frac{
                \mathbf{S}\overline{\mathbf{A}}_i^{\top}
            }{
                \sqrt{d_q}
            }
        \right)
        \overline{\mathbf{A}}_i
    \right],
    \label{eq:task_slot_construction}
\end{equation}
where the softmax is applied row-wise over the anchors, $\mathbf{S}\in\mathbb{R}^{K\times d_q}$ contains $K$ learnable slot vectors shared across tasks, and $\mathbf{Z}_i=[\mathbf{z}_{i,1},\ldots,\mathbf{z}_{i,K}]^{\top}$ contains the task slots derived from $\mathcal{A}_i$. Thus, all tasks share the same slot construction, while the content of $\mathbf{Z}_i$ is determined by the corresponding anchor set $\mathcal{A}_i$.
The normalized mean of the task slots defines a coarse address for task $i$:
\begin{equation}
    \mathbf{T}_c^i
    =
    \mathrm{Norm}
    \left(
        \frac{1}{K}
        \sum_{k=1}^{K}
        \mathbf{z}_{i,k}
    \right).
\end{equation}
Given a sample query $\mathbf{q}$, TAILS first evaluates the relevance of each accumulated task structure and then the relevance of its individual task slots:
\begin{equation}
    \pi_i(\mathbf{q})
    =
    \mathrm{softmax}_{i}
    \left(
        \mathbf{q}^{\top}\mathbf{T}_c^i
    \right),
\end{equation}
\begin{equation}
    \rho_{i,k}(\mathbf{q})
    =
    \mathrm{softmax}_{k}
    \left(
        \mathbf{q}^{\top}\mathbf{z}_{i,k}
    \right).
\end{equation}
\begin{equation}
\label{eq:latent_recall}
    \mathbf{r}
    =
    \sum_{i=1}^{t}
    \pi_i(\mathbf{q})
    \sum_{k=1}^{K}
    \rho_{i,k}(\mathbf{q})
    \mathbf{z}_{i,k}.
\end{equation}
We refer to $\mathbf{r}$ as \emph{latent recall}: an input-conditioned summary of accumulated feature structures that provides auxiliary evidence for interpreting the current representation. The task organization acts only as an indexing scaffold rather than a hard task decision. Since $\pi_i(\mathbf{q})$ is soft, latent recall may combine relevant task slots from multiple tasks when semantic neighborhoods cross the procedural boundaries of continual training. The resulting $\mathbf{r}$ provides the historical context that TAILS internalizes through a conditioned feature displacement. Specifically, the current feature determines a candidate correction direction, while latent recall modulates which dimensions of that correction are expressed:
\begin{equation}
\label{eq:TAILSoverall}
    \mathbf{h}'
    =
    \mathcal{F}_{\mathrm{TAILS}}(\mathbf{h};\mathcal{A})
    =
    \mathrm{Norm}\!\left(
        \mathrm{LN}(W_b\mathbf{h})
        +
        \Delta\mathbf{h}(\mathbf{h},\mathbf{r})
    \right),
\end{equation}
\begin{equation}
    \Delta\mathbf{h}(\mathbf{h},\mathbf{r})
    =
    W_o\mathrm{GELU}\!\left(
        W_h\mathrm{LN}(\mathbf{h})
        \odot
        \sigma\!\left(W_r\mathrm{LN}(\mathbf{r})\right)
    \right).
\end{equation}
Here, $W_b$ is an identity-initialized linear mapping, allowing this branch to start from the original PTM feature and avoid disturbing pre-trained semantics at the beginning of TAILS training \cite{houlsby2019parameter}.
Unlike conventional residual feature adaptation conditioned only on the current representation, TAILS conditions its correction on latent recall derived from accumulated task structures.


\subsection{Latent-Recall-Preserving Optimization}
The TAILS module is incrementally updated after each task. Once the base learner completes task $t$, we freeze its representation learner and classifier and optimize only TAILS in the feature space. Let $\mathcal{F}_{t}^{\mathrm{cur}}$ denote the set of post-update representations extracted from the current-task samples. The optimization uses $\mathcal{F}_{t}^{\mathrm{cur}}$ together with the historical anchor set $\mathcal{A}_{1:t-1}=\bigcup_{i=1}^{t-1}\mathcal{A}_i$. These two sources play complementary roles: current-task representations incorporate newly acquired structures into the shared TAILS mapping, whereas historical anchors provide stable reference features for preserving previously established classifier compatibility, recall addresses, and local feature geometry without retaining any previous training samples. Since the anchors remain fixed and the base learner is never revisited, this procedure updates only how TAILS interprets the accumulated feature space without modifying the original continual-learning model. Accordingly, each training feature
$\mathbf{h}_{\mathrm{train}}\in
\mathcal{F}_{t}^{\mathrm{cur}}\cup\mathcal{A}_{1:t-1}$
is shaped as
$\mathbf{h}'_{\mathrm{train}}
=
\mathcal{F}_{\mathrm{TAILS}}
(\mathbf{h}_{\mathrm{train}};\mathcal{A})$,
and its corresponding query $\mathbf{q}_{\mathrm{train}}$ is computed from
$\mathbf{h}_{\mathrm{train}}$ using Eq.~\eqref{eq:tails_query}.
We impose three complementary constraints: 
\begin{equation}
    \mathcal{L}_{\mathrm{TAILS}}
    =
    \mathcal{L}_{\mathrm{cls}}
    +
    \mathcal{L}_{\mathrm{task}}
    +
    \mathcal{L}_{\mathrm{stab}}.
\end{equation}
The classifier-compatibility constraint ensures that the shaped feature remains meaningful under the frozen decision rule of base learner $\mathcal{B}$:
\begin{equation}
    \mathcal{L}_{\mathrm{cls}}
    =
    \mathrm{CE}\!\left(
        G_{\mathcal{B}}(\mathbf{h}'_{\mathrm{train}}),y
    \right).
\end{equation}
Gradients pass through the fixed classifier only to TAILS. Thus, the objective does not learn a new classifier; it teaches latent recall to produce feature corrections that are compatible with the base learner's established semantic space.

The recall-addressing constraint organizes where the current query should seek evidence. Let $\tau(y)$ denote the task ID associated with the class $y$. We supervise the similarities between the query and accumulated task centers:
\begin{equation}
    \mathcal{L}_{\mathrm{task}}
    =
    \mathrm{CE}\!\left(
        \left\{
        \mathrm{sim}_{\cos}
        \left(\mathbf{q}_{\mathrm{train}},\mathbf{T}_c^i\right)
        \right\}_{i=1}^{t},
        \tau(y)
    \right).
\end{equation}
This signal learns the addressing behavior of latent recall without requiring a task ID at inference. 
The supervised center response establishes a reliable address, whereas Eq.~\eqref{eq:latent_recall} still performs soft composition of task slots for the actual feature correction.

The semantic-preservation constraint limits excessive deviation from the original PTM representation:
\begin{equation}
    \mathcal{L}_{\mathrm{stab}}
    =
    1-
    \frac{
        \mathbf{h}'_{\mathrm{train}}{}^{\top}
        \mathbf{h}_{\mathrm{train}}
    }
    {
        \|\mathbf{h}'_{\mathrm{train}}\|_2
        \|\mathbf{h}_{\mathrm{train}}\|_2
    }.
\end{equation}
Together, the three terms preserve classifier compatibility, recall-address consistency, and continuity of PTM semantics, allowing TAILS to incorporate new task structures while maintaining a coherent latent-recall across all tasks.

\subsection{Integrating TAILS with PTM Learners}

TAILS operates at the representation interface shared by otherwise heterogeneous PTM-based learners. This placement allows the same unified module to assist inference while preserving each base learner's adaptation mechanism and original decision rule.

For methods that ultimately produce a single representation for prediction, including raw PTM, prompt-based methods, and adapter-based methods that fuse multiple adapter outputs into one \cite{wang2025tuna}, TAILS applies $\mathcal{F}_{\mathrm{TAILS}}(\cdot;\mathcal{A})$ to the resulting feature $\mathbf{h}$ and feeds the shaped representation $\mathbf{h}'$ from Eq.~\eqref{eq:TAILSoverall} to the original classifier. For methods that expose task-specific candidate features before prediction, such as task-specific adapters \cite{sun2025mos} or task-specific LoRA branches \cite{he2025cllora}, the same TAILS module is applied independently to each candidate feature:
\begin{equation}
    \mathbf{h}'^{\,i}
    =
    \mathcal{F}_{\mathrm{TAILS}}
    \left(
        \mathbf{h}^{i};\mathcal{A}
    \right),
    \quad i=1,\cdots,t.
\end{equation}
The base learner then follows its original candidate selection, aggregation, or prediction procedure. Thus, every candidate is interpreted through the same accumulated recall space without changing the semantics of the base method.

TAILS is also compatible with different classifier types. For a parametric classifier, such as a linear head, the shaped representation is passed directly to the original classifier:
\begin{equation}
    \hat{y}
    =
    \arg\max_c
    G_{\mathcal{B}}
    \left(
        \mathbf{h}'
    \right)_c .
\end{equation}
For prototype-based cosine classifiers \cite{sun2025mos,zhou2025revisiting}, the classifier is parameterized by stored class prototypes. Since TAILS changes the representation used for prediction, each class prototype is transformed by the same unified mapping to preserve sample--prototype consistency:
\begin{equation}
    \boldsymbol{\mu}_c'
    =
    \mathcal{F}_{\mathrm{TAILS}}
    \left(
        \boldsymbol{\mu}_c;\mathcal{A}
    \right).
\end{equation}
The prediction rule is then:
\begin{equation}
    \hat{y}
    =
    \arg\max_c
    \mathrm{sim}_{\cos}
    \left(
        \mathbf{h}', \boldsymbol{\mu}_c'
    \right).
\end{equation}
Consequently, deploying TAILS requires no structural modification to the original method. Its prediction mechanism remains intact, while task-anchored latent recall only reshapes the representation supplied to it.

\newcolumntype{Y}{>{\centering\arraybackslash}X}
\begin{table*}[!t]
    \centering

    \definecolor{rawptmcolor}{RGB}{226,246,232}
    \definecolor{promptcolor}{RGB}{226,246,232}
    \definecolor{adaptercolor}{RGB}{236,224,245}
    \definecolor{loracolor}{RGB}{250,246,210}

    \def\TAILSTableWidth{0.98\textwidth}
    \def\TAILSTableScale{1.0}
    \def\myrowcolor#1{\rowcolor{#1}[2.0\tabcolsep][2.0\tabcolsep]}
    \def\dataset#1#2{\shortstack[c]{#1\\[-0.5pt]{\small #2}}}
    \newcommand{\score}[2]{\mbox{$#1{\scriptscriptstyle\pm #2}$}}
    \newcommand{\bestscore}[2]{\mbox{$\mathbf{#1}{\scriptscriptstyle\pm #2}$}}

    \setlength{\tabcolsep}{2.6pt}
    \renewcommand{\arraystretch}{1.08}

    \makebox[\textwidth][c]{%
    \scalebox{\TAILSTableScale}{%
    \begin{minipage}{\TAILSTableWidth}
    \scalebox{1}{
    \begin{tabularx}{\linewidth}{@{}lYYYYYYYY@{}}
    \toprule
    \multirow{2}{*}{Method}
    & \multicolumn{2}{c}{\dataset{CARS-196}{B16 Inc20}}
    & \multicolumn{2}{c}{\dataset{ImageNetA-200}{B20 Inc20}}
    & \multicolumn{2}{c}{\dataset{ImageNetR-200}{B5 Inc5}}
    & \multicolumn{2}{c}{\dataset{VTAB-50}{B10 Inc10}} \\
    \cmidrule(lr){2-3}
    \cmidrule(lr){4-5}
    \cmidrule(lr){6-7}
    \cmidrule(lr){8-9}
    & $A^{\text{task}}_T$ & $A_T$
    & $A^{\text{task}}_T$ & $A_T$
    & $A^{\text{task}}_T$ & $A_T$
    & $A^{\text{task}}_T$ & $A_T$ \\
    \midrule

    \myrowcolor{rawptmcolor}
    SimpleCIL
    & \score{43.53}{0.01} & \score{37.68}{0.01}
    & \score{52.98}{0.01} & \score{48.67}{0.01}
    & \score{55.19}{0.01} & \score{54.59}{0.02}
    & \score{87.11}{0.01} & \score{84.33}{0.01} \\

    \myrowcolor{rawptmcolor}
    SimpleCIL-TAILS
    & \score{68.03}{0.97} & \score{64.39}{0.95}
    & \score{58.70}{0.66} & \score{54.86}{0.60}
    & \score{68.69}{0.99} & \score{67.94}{0.93}
    & \score{91.98}{0.08} & \score{90.18}{0.04} \\

    \myrowcolor{promptcolor}
    DualPrompt
    & \score{34.90}{0.92} & \score{32.73}{0.96}
    & \score{59.48}{0.80} & \score{56.71}{0.84}
    & \score{65.69}{0.94} & \score{65.49}{0.98}
    & \score{79.75}{2.84} & \score{79.07}{2.90} \\

    \myrowcolor{promptcolor}
    DualPrompt-TAILS
    & \score{61.70}{0.95} & \score{59.76}{0.93}
    & \score{65.07}{0.83} & \score{60.86}{0.81}
    & \score{74.30}{0.95} & \score{73.89}{0.97}
    & \score{90.16}{2.85} & \score{88.30}{2.89} \\

    \myrowcolor{adaptercolor}
    SEMA
    & \score{57.05}{0.85} & \score{55.54}{0.91}
    & \score{64.82}{0.35} & \score{58.62}{0.41}
    & \score{69.31}{0.85} & \score{69.11}{0.91}
    & \score{94.93}{0.86} & \score{89.51}{0.92} \\

    \myrowcolor{adaptercolor}
    SEMA-TAILS
    & \score{67.23}{0.90} & \score{65.72}{0.86}
    & \score{67.71}{0.40} & \score{61.12}{0.36}
    & \score{70.82}{0.86} & \score{70.56}{0.90}
    & \score{98.48}{0.88} & \score{92.03}{0.90} \\

    \myrowcolor{adaptercolor}
    MOS
    & \score{73.52}{0.87} & \score{70.66}{0.89}
    & \score{58.81}{0.39} & \score{55.20}{0.37}
    & \score{71.17}{0.89} & \score{70.55}{0.87}
    & \score{99.54}{0.91} & \score{92.25}{0.87} \\

    \myrowcolor{adaptercolor}
    MOS-TAILS
    & \score{75.48}{0.89} & \score{72.75}{0.87}
    & \score{60.92}{0.36} & \score{57.96}{0.40}
    & \score{73.90}{0.85} & \score{73.32}{0.91}
    & \bestscore{99.65}{0.85} & \score{92.94}{0.93} \\

    \myrowcolor{adaptercolor}
    TUNA
    & \score{77.66}{0.90} & \score{76.61}{0.86}
    & \score{66.21}{0.37} & \score{64.18}{0.39}
    & \score{75.74}{0.87} & \score{75.32}{0.89}
    & \score{98.28}{0.90} & \score{91.80}{0.88} \\

    \myrowcolor{adaptercolor}
    TUNA-TAILS
    & \bestscore{79.06}{0.87} & \bestscore{78.09}{0.89}
    & \bestscore{67.73}{0.41} & \bestscore{65.30}{0.35}
    & \bestscore{77.16}{0.89} & \bestscore{76.74}{0.87}
    & \score{98.44}{0.91} & \score{92.27}{0.87} \\

    \myrowcolor{loracolor}
    CL-LoRA
    & \score{65.49}{0.51} & \score{64.36}{0.55}
    & \score{59.41}{0.12} & \score{57.62}{0.16}
    & \score{73.62}{0.61} & \score{73.49}{0.65}
    & \score{97.67}{0.32} & \score{94.22}{0.36} \\

    \myrowcolor{loracolor}
    CL-LoRA-TAILS
    & \score{70.12}{0.53} & \score{69.41}{0.56}
    & \score{61.06}{0.13} & \score{59.33}{0.15}
    & \score{74.15}{0.66} & \score{74.01}{0.62}
    & \score{98.18}{0.33} & \bestscore{94.71}{0.35} \\

    \bottomrule
    \end{tabularx}
    }
    \end{minipage}
    }}
    \caption{Comparison of task prediction accuracy $A^{\text{task}}_T$ and final classification accuracy $A_T$. ImageNet-R uses the B5 Inc5 split with 200 classes, resulting in a 40-task long-sequence benchmark. The best results are highlighted in bold.}
    \label{tab:ptm_TAILS_main}
\end{table*}
\section{Experiments}
This section conducts extensive experiments to evaluate TAILS. 
Sec.~\ref{subsec:benchmark} compares TAILS with representative PTM-based CL methods across different adaptation paradigms. 
Sec.~\ref{subsec:abl} analyzes the training objectives, underlying mechanism, inference efficiency, and parameter overhead. 
Sec.~\ref{subsec:sensitivity} studies the sensitivity of TAILS to the number of support anchors and task slots. 
T-SNE visualizations of the feature spaces are provided in the supplementary material.

\subsection{Implementation Details}
\label{subsec:5.1}
\textbf{Dataset and splits.} We select four benchmark datasets that exhibit notable domain gaps \cite{zhou2024ease} relative to the ImageNet-21K pretraining distribution, including Stanford Cars \cite{krause2013cars}, ImageNet-A \cite{hendrycks2021natural}, ImageNet-R \cite{hendrycks2021many}, and VTAB \cite{zhai2019large}.
Following the ``B/Base-m, Inc-n'' protocol \cite{zhou2025revisiting}, datasets are split into CARS B16 Inc20, ImageNet-A B20 Inc20, VTAB B10 Inc10, and ImageNet-R B5 Inc5. ImageNet-R with B5 Inc5 forms a 40-task benchmark, allowing us to evaluate whether TAILS remains effective when task-level ambiguity accumulates over long-sequence CL. 

\noindent\textbf{Baselines.} We deploy TAILS on 
baseline or SOTA PTM-based CL approaches that cover different paradigms: prompt-based method DualPrompt \cite{wang2022dualprompt}, raw PTM and prototype classification method SimpleCIL \cite{zhou2025revisiting}, multi-task adapter methods MOS \cite{sun2025mos} and SEMA \cite{wang2025sema}, adapter-based integration method TUNA \cite{wang2025tuna}, and task-specific LoRA method CL-LoRA \cite{he2025cllora}.
Following the original settings of each method, DualPrompt and SEMA use ViT-B/16-224, while the remaining methods adopt ViT-B/16-IN21K as the PTM backbone.

\noindent\textbf{Programming and hyperparameters.} All experiments are implemented in PyTorch 2.4.1 and conducted on NVIDIA 3090 GPUs. Baseline settings follow the configurations in original papers or LAMDA-Pilot \cite{sun2023pilot}.
For TAILS, the number of task slots $K$ is mainly set to 5, and the number of support anchors for each class is mainly set to 20. TAILS adopts the training configuration of the corresponding base method, including the optimizer, initial learning rate, batch size, number of epochs, and learning-rate schedule.

\noindent\textbf{Evaluation metrics.} We use $A_T$ to evaluate the final classification accuracy on the joint test set of all tasks after training on the last task, and use $A^{\text{task}}_T$ to measure whether the task of the predicted class matches the ground-truth task.

\subsection{Benchmark Comparison}
\label{subsec:benchmark}
\begin{figure}[!t]
  \centering
  \subfloat[ImageNet-A]{
    \includegraphics[height=1.35in,keepaspectratio]{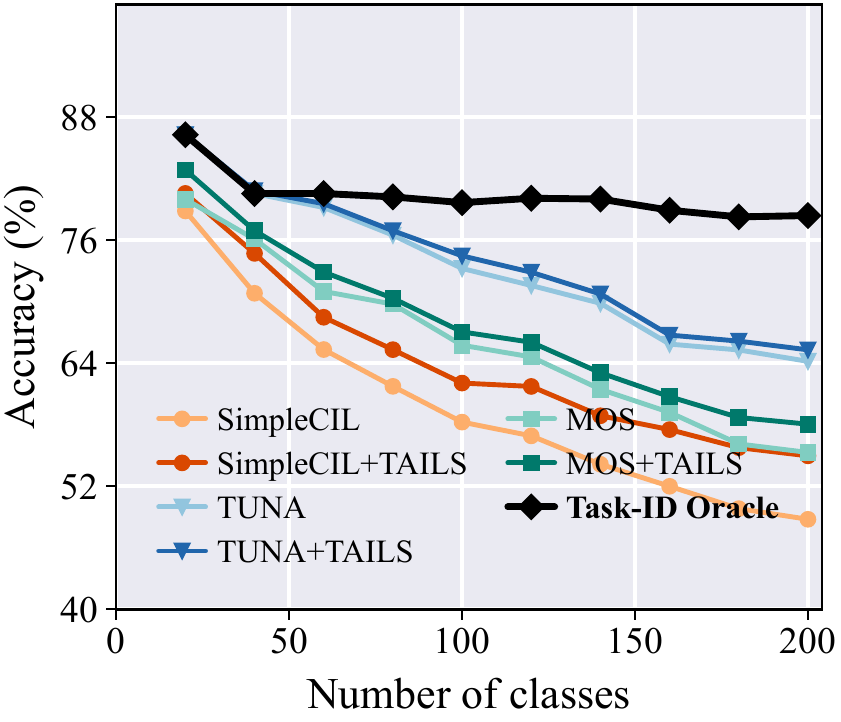}
    }  
  \subfloat[ImageNet-R]{
    \includegraphics[height=1.35in,keepaspectratio]{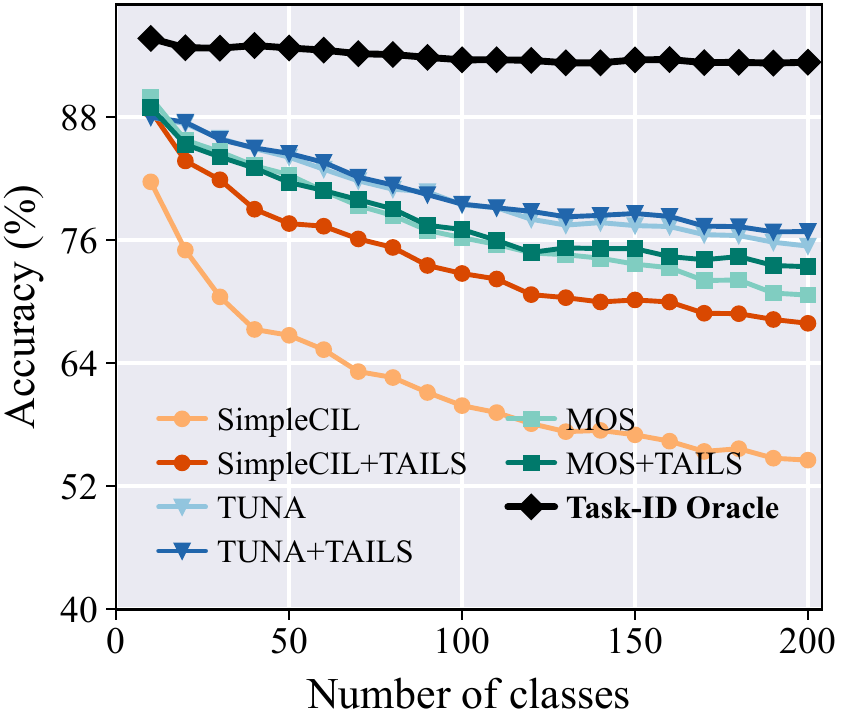}
    }  
  \caption{Accuracy trajectories of different methods on ImageNet-A and ImageNet-R.}
  \label{fig:curve1}
\end{figure}

This subsection evaluates TAILS by deploying it on various PTM-based CL methods.
Table~\ref{tab:ptm_TAILS_main} reports the mean and standard deviation of $A^{\text{task}}_T$ and $A_T$ over five seeds on four benchmark datasets. 
For high-performing baselines, $A^{\text{task}}_T$ is often close to $A_T$, indicating that their within-task prediction is already strong, while cross-task inference becomes the main factor limiting final performance.
TAILS effectively improves both cross-task prediction accuracy and final classification accuracy, enabling the enhanced learners to surpass existing state-of-the-art PTM-based CL methods on multiple benchmarks.
The improvements are also consistent on the 40-task ImageNet-R dataset, demonstrating that TAILS remains effective on the long-sequence CL benchmark.
These results demonstrate the broad applicability of TAILS to diverse PTM-based CL paradigms.
In particular, TAILS still yields notable improvements when applied to strong SOTA baselines such as TUNA and CL-LoRA. 
We further report the accuracy trajectories throughout continual learning in Fig.~\ref{fig:curve1}. TAILS consistently performs best in all evaluated settings, and the gains remain evident as more classes are introduced. The complete curves for all methods and datasets are provided in the supplementary material.

\begin{table}[!ht]
    \centering
    \resizebox{0.95\columnwidth}{!}{
    \begin{tabular}{ccc|cc}
    \bottomrule
    $\mathcal{L}_{\mathrm{cls}}$ & $\mathcal{L}_{\mathrm{task}}$ & $\mathcal{L}_{\mathrm{stab}}$ & SimpleCIL-TAILS & TUNA-TAILS \\ \hline
    $\times$ & $\times$ & $\times$ & 48.67 & 64.18 \\
    $\times$ & \checkmark & \checkmark & 50.51 & 65.17 \\
    \checkmark & $\times$ & \checkmark & 54.12 & 64.91 \\
    \checkmark & \checkmark & $\times$ & 54.78 & 65.02 \\
    \checkmark & \checkmark & \checkmark & \textbf{54.86} & \textbf{65.30} \\
    \bottomrule
    \end{tabular}
    }
    \caption{Ablation study of the training objectives in TAILS. All results are averaged over five seeds.}
    \label{tab:abl}
\end{table}
\begin{table}[!t]
    \centering

    \setlength{\tabcolsep}{2.0pt}
    \renewcommand{\arraystretch}{1.04}

    \resizebox{1\columnwidth}{!}{%
    \begin{tabular}{lcccccccc}
        \toprule
        \multirow{2}{*}{Variant}
        & \multicolumn{4}{c}{CARS}
        & \multicolumn{4}{c}{ImageNet-R} \\
        \cmidrule(lr){2-5}
        \cmidrule(lr){6-9}
        & SimpleCIL & TUNA & MOS & CL-LoRA
        & SimpleCIL & TUNA & MOS & CL-LoRA \\
        \midrule

        Anchor-MLP
        & 58.81 & 77.32 & 71.34 & 65.92
        & 62.43 & 75.91 & 72.15 & 73.57 \\

        Anchor Retrieval
        & 56.66 & 77.04 & 71.03 & 66.02
        & 60.47 & 75.51 & 71.37 & 73.48 \\

        Single-Center
        & 62.23 & 77.72 & 70.54 & 67.52
        & 65.95 & 76.23 & 72.55 & 73.91 \\

        Logit Calibration
        & 45.22 & 76.63 & 70.81 & 64.61
        & 56.82 & 75.33 & 70.42 & 73.31 \\

        \midrule

        TAILS
        & \textbf{64.39}
        & \textbf{78.09}
        & \textbf{72.75}
        & \textbf{69.41}
        & \textbf{67.94}
        & \textbf{76.74}
        & \textbf{73.32}
        & \textbf{74.01} \\

        \bottomrule
    \end{tabular}%
    }
    \caption{Mechanism comparison on CARS and ImageNet-R.
    All results are averaged over five seeds.
    All variants use the same anchor and optimization budgets.}
    \label{tab:mechanism}
\end{table}
\subsection{Ablation Study}
\label{subsec:abl}
This subsection conducts a series of ablation studies on TAILS. 
We first investigate the effectiveness of the three training objectives in TAILS on the ImageNetA dataset, as shown in Table~\ref{tab:abl}. 
The results show that $\mathcal{L}_{\mathrm{cls}}$, $\mathcal{L}_{\mathrm{task}}$, and $\mathcal{L}_{\mathrm{stab}}$ all contribute to the final performance. 
The relative importance of $\mathcal{L}_{\mathrm{cls}}$ and $\mathcal{L}_{\mathrm{task}}$ varies across different base learners. 
For methods such as SimpleCIL, where classification is not sufficiently optimized during continual learning, $\mathcal{L}_{\mathrm{cls}}$ plays a more important role in directly improving classification accuracy. 
In contrast, TUNA has already learned task-specific classifiers through task-wise adaptation, so $\mathcal{L}_{\mathrm{task}}$ becomes critical for organizing task-level responses and improving task inference. 
$\mathcal{L}_{\mathrm{stab}}$ further provides a consistent gain by preventing the shaped representation from drifting excessively away from the original PTM feature.

Moreover, we compare TAILS with four competitive variants under the same anchor and optimization budgets, as shown in Table ~\ref{tab:mechanism}. All results are averaged over five random seeds. For each seed, the same sampled anchors are shared
by TAILS and all mechanism variants. Anchor-MLP removes the TAILS architecture and uses anchors only to assist the incremental update of an MLP. Anchor Retrieval directly computes attention between the input feature and all stored anchors, and uses their weighted sum for feature shaping. Single-Center just represents each task with one center rather than multiple slots. Logit Calibration applies TAILS to the logits instead of feature shaping. TAILS consistently performs best across all evaluated settings. The consistent gains over Anchor-MLP, Anchor Retrieval, and Single-Center validate the TAILS module design. Direct logit calibration is brittle because even small corrections can change class rankings and destabilize the final prediction. Moreover, cross-task ambiguity is a representation-level problem that is difficult to resolve by modifying the classifier alone.

\begin{table}[t]
    \centering
    \scalebox{0.95}{
    \begin{tabular}{ccccc}
    \toprule
    \multirow{2}{*}{Method}
    & \multicolumn{2}{c}{ImageNet-A}
    & \multicolumn{2}{c}{ImageNet-R} \\
    \cmidrule(r){2-3} \cmidrule(r){4-5}
    & $A_T$ & $\bar{t}_{\mathrm{inf}}$ (ms)
    & $A_T$ & $\bar{t}_{\mathrm{inf}}$ (ms)\\
    \midrule
    SimpleCIL
    & 48.67 & 4.73
    & 54.59  & 4.82 \\
    SimpleCIL-TAILS
    & 54.86 & 4.92
    & 67.94 & 5.10 \\
    ACMap
    & 54.49 & 4.93
    & 69.33 & 5.05 \\
    TUNA
    & 64.18 & 62.78
    & 75.32  & 145.04 \\
    TUNA-TAILS
    & 65.30 & 66.09
    & 76.74 & 150.42 \\
    \bottomrule
    \end{tabular}
    }
    \caption{Comparison of average final accuracy and the per-sample inference time measured over the test set.}
    \label{tab:inf}
\end{table}

Furthermore, we investigate the inference-time overhead introduced by TAILS. 
For a more comprehensive comparison, we include ACMap \cite{Fukuda_2025_CVPR}, a PTM-based method specifically optimized for inference speed as an efficiency-oriented reference.
We report the mean final accuracy and the average per-sample inference time on the test set over multiple runs with five seeds, as shown in Table~\ref{tab:inf}.
TAILS introduces only marginal inference overhead over the original method. 
Notably, applying TAILS to SimpleCIL directly yields a learner that is competitive with state-of-the-art efficiency-oriented methods in both accuracy and inference speed.

\begin{table}[!ht]
    \centering
    \resizebox{0.95\columnwidth}{!}{
    \begin{tabular}{lccc}
    \toprule
    Methods & Model Params. & Auxiliary Storage & Total Storage \\
    \midrule
    ACMap   & 1.19  & 47.59  & 48.78 \\
    MOS     & 12.17 & 130.29 & 142.46 \\
    TUNA    & 12.48 & 118.12 & 130.60 \\
    CL-LoRA & 7.56  & 0.00   & 7.56 \\
    TAILS   & 1.14  & 3.07   & 4.21 \\
    \bottomrule
    \end{tabular}
    }
    \caption{Comparison of additional storage overhead on the ImageNet-R dataset, measured in millions of scalar values (M), excluding the PTM.}
    \label{tab:param}
\end{table}
In addition, Table~\ref{tab:param} compares the additional storage required by different methods under the ImageNet-R protocol. For a consistent comparison, we count both the model parameters and all auxiliary tensors that must be retained for subsequent learning or inference. The auxiliary storage includes the task anchors maintained by TAILS, the class-alignment statistics used by MOS and TUNA, and the task-specific adapter checkpoints retained by ACMap. For CL-LoRA, all task-expanded LoRA weights are included in the model parameters, with no separately stored auxiliary tensors. TAILS is lightweight on its own, and its plug-in overhead remains below the learner-specific storage introduced by every compared method. Relative to the storage of the corresponding base learner, adding TAILS incurs only 3.0\% overhead for MOS, 3.2\% for TUNA, and 8.6\% for ACMap; even for the more compact CL-LoRA, the additional storage remains below its original task-expanded parameter footprint.

\subsection{Sensitivity Analysis of Hyperparameters}
\label{subsec:sensitivity}
\begin{figure}[!ht]
  \centering
  \subfloat[Support Anchors]{
    \includegraphics[height=1.37in,keepaspectratio]{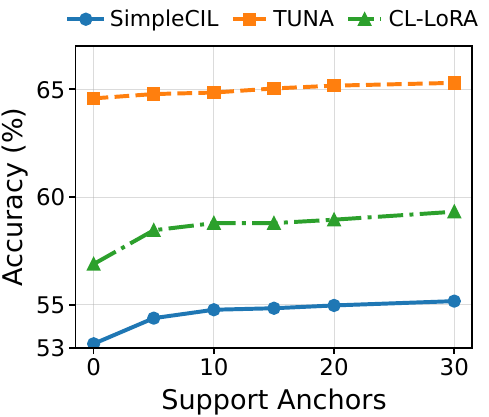}
    \label{fig:figsa}
    }
  \subfloat[Task Slots]{
    \includegraphics[height=1.37in,keepaspectratio]{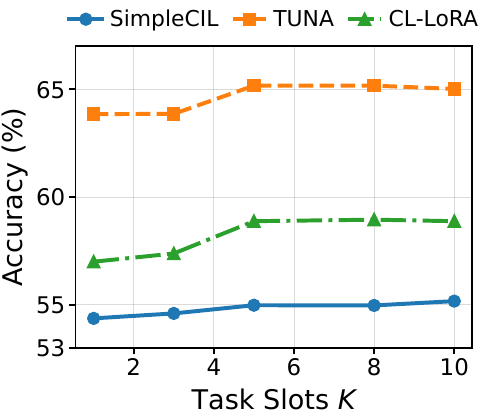}
    \label{fig:figts}
    }  
  \caption{Impact of support anchor and task slot counts on the results across different PTM-based CL methods.}
  \label{fig:figparams}
\end{figure}

We study how key hyperparameters in TAILS affect performance, with the results shown in Fig.~\ref{fig:figparams}. 
The results indicate that even when no support anchors are used, TAILS can still improve the base learner by relying solely on class prototypes as task anchors. 
Increasing the number of support anchors generally improves final classification accuracy, as additional anchors capture more within-class variation and support richer task-slot representations for composing latent recall.
However, the gains diminish as additional anchors become increasingly redundant.
Increasing the number of task slots further improves performance by capturing finer structures within each task. Too many slots, however, can fragment the task representation and make query-slot matching less reliable, thereby degrading latent-recall composition.

\section{Conclusion}
This paper introduces TAILS, a task-anchored latent-recall-assisted inference module for PTM-based continual learning. TAILS interprets each current representation against fixed accumulated task structures, composes input-conditioned latent recall, and internalizes the recalled evidence as a recall-conditioned feature correction.
It is shared across all learned tasks and incrementally updated in a decoupled manner, so accumulated task-level information can refine current representations while preserving the original training pipeline of the base learner. We further store class prototypes and support features as fixed task anchors in the feature space, providing stable references for training and inference. Extensive experiments across various continual learning methods demonstrate that TAILS consistently improves baseline performance with modest parameter overhead and negligible inference cost.
Future work will explore more expressive anchor construction strategies and stronger shaping objectives to further enhance task-level discriminability.

\bibliography{main}

\end{document}